\documentclass[11pt,a4paper]{article}
\usepackage[hyperref]{acl2018}
\usepackage{times}
\usepackage{graphicx}
\usepackage{appendix}

\usepackage{url}

\aclfinalcopy 

\title{Sentiment Analysis for Open Domain Conversational Agent}

\author{Mohamad Alissa, Issa Haddad, Jonathan Meyer, Jade Obeid, \\
  \textbf{Nicolas Wiecek, Sukrit Wongariyakavee} \\
  Conversational Agents, Department of Computer Science \\
  Heriot-Watt University \\
  Edinburgh, UK \\
  {\tt \{mmm18, ih19, jcm5, jo20, kv10, nw86, sw83\}@hw.ac.uk} 
  \\}

\date{}

\begin{document}
\maketitle
\begin{abstract}
  
  The applicability of common sentiment analysis models to open domain human robot interaction is investigated within this paper. The models are used on a dataset specific to user interaction with the Alana system (a Alexa prize system) in order to determine which would be more appropriate for the task of identifying sentiment when a user interacts with a non-human driven socialbot. With the identification of a model, various improvements are attempted and detailed prior to integration into the Alana system. The study showed that a Random Forest Model with 25 trees trained on the dataset specific to user interaction with the Alana system combined with the dataset present in NLTK Vader outperforms other models. The new system (called 'Rob') matches it's output utterance sentiment with the user's utterance sentiment. This method is expected to improve user experience because it builds upon the overall sentiment detection which makes it seem that new system sympathises with user feelings. Furthermore, the results obtained from the user feedback confirms our expectation.
  
\end{abstract}

\section{Introduction}

    The aim of sentiment analysis is to extract subjective information like opinion and sentiment from text in natural language to create actionable knowledge to be used by a decision support system \cite{Pozzi20171}. Organisations and individuals seek to know the opinions of others to form better decisions, this can be seen in consumer data information, the use of focus groups, advert placement in internet sidebars, opinionated documents, civilian response in crisis situations and the medical field \cite{Pozzi20171}. Sentiment analysis continues to be highly challenging with the research community attempting many sub-problems that have not been completely solved \cite{Pozzi2017239}. With this in mind, it is expected that scripted conversations between two humans like what is done in movies, unscripted conversations between two humans, and human-machine interaction systems will contain a varying amount of sentiment with very different dialogue. \\
    \indent Working with a large dataset in the area of human-machine interaction systems allows the evaluation of already existing tools and machine learning techniques to better optimise development within this area. The model is integrated into Alana (a 2017 Alexa prize system \cite{RamAshwin2018CATS} consisting of an ensemble of bots, combining rule-based and machine learning systems, and using a contextual ranking mechanism to choose system responses) \cite{PapaioannouIoannis2017AEMw}. In order to evaluate the improvements to the system (Rob) a lightweight design is required with minimal interference to the original system (Alana).

\section{Related Work}

\indent In the original Alana system that acts as the base that is improved upon, \citet{PapaioannouIoannis2017AEMw} used sentiment polarity as weighting for the Linear Classifier Ranker which ranked Bucket responses. This allowed the highest-scoring response to be selected as the output utterance. While the linear ranker of the system was later replaced with a neural model at the finals of the Amazon Alexa Challenge, improvements to the sentiment polarity could see a marked increase in responses that would improve user experience. \\
\indent \citet{Singh2016} explore techniques for opinion extraction from huge amount of unstructured information with recent papers on sentiment analysis detailing the techniques used as well as the dataset used. With over 60k dialogue instances available in the dataset to be used identifying complexity of techniques and their speed allows better judgement over the models that should be looked at. \\
\indent Sentiment is linked to previous context, \citet{KATZ2015162} details a way to retrieve key terms in the text and analyse the context in which they appear. The detected terms are used to generate features for supervised learning, the advantage of this approach is incorporation of context that works well over multiple domains with varying degrees of noise. \\
\indent Another approach presented by \citet{APPEL2016110} uses a hybrid system that works well at the sentence level against twitter-like datasets using sentiment lexicons, NLP essential techniques and fuzzy sets. 

\section{System Design, Architecture, and Components}

\subsection{Dataset and Annotation}

\subsubsection{Interactions dataset}

In order to train and evaluate the models used in this work, a dataset of human-machine dialogues was used. To the best of our knowledge, very few pre-annotated datasets on this type of interaction are currently available. As such, and to ensure the best level of performance, a dataset of transcribed dialogues was manually annotated. This data originates from conversations between American citizens and the Alana system in 2017, through the Alexa voice assistant.  
The annotation scheme used consists of two polarities (i.e. positive and negative), two intensity levels and a neutral sentiment. The resulting annotation scheme is ['Very positive', 'Positive', 'Neutral', 'Negative', 'Very Negative'], also written as ('++', '+', '0','-', '--'). The annotation is done solely on the user utterances, without context; a possibility that was explored but which would result in complex models which would be capable to handle two or more utterances at a time.
To ensure high inter-annotator agreement, sample utterances where collectively annotated, especially to agree on what is considered to be of the strongest intensity level.
A dataset of 1200 utterances was selected from the main Alana interactions dataset either randomly or due to the presence of a 'sentiment-related' word in the sentence. Hu and Liu's opinion lexicon, available at https://www.cs.uic.edu/~liub/FBS/sentiment-analysis.html, was used to identify such sentences. 
These 1400 utterances where split into 7 equal size parts (for each members), with a 10\% overlap between datasets, meaning that every member annotated 200 uniques utterances along with 20 utterances from an other annotator. This overlap was used to compute the Kappa value (inter-annotator agreement metric). The annotators has the possibility to skip utterances they believed were not of good quality (such as a randomly selected sentence : "Ah") or for which they were unsure due to possible sarcasm or strong context dependency. 
If we ignore utterances that were skipped by one annotator, the overall Kappa value is 0.764, otherwise \footnote{Meaning that if an annotator skipped an utterance while the other didn't, this will result in a non-agreement.}, the Kappa value is equal to 0.6999. Note that a Kappa value above 0.7 is considered a reflection of a satisfactory inter-annotator agreement.
The final dataset consists of 1164 annotated utterances.

\subsubsection{Sentiment words dataset}
Due to the limited amount of annotated data, we quickly realised that the trained models would fail to capture some clearly sentimental sentences such as "I am confused", this is due to the absence of some words in the annotated dataset.
In order to address this problem, at least partially, a secondary dataset was used along with the annotated one. This dataset originates from the Vader lexicon where a collection of words have a sentiment value associated with them (floating value between -4.0 (strongly negative) and +4.0 (strongly positive)).
We chose not to use the full dataset as it consisted of too many training samples (compared to our annotated dataset). Instead, we selected the "strongest" sentiment words (absolute sentiment value greater than 2.5). The real-valued sentiment were discretized using the following scheme : ['Very negative' : [-4.0, -3.01], 'Negative' : [-3.0, -2.5], Neutral : [], 'Positive':[2.5, 3.0], 'Very positive : [3.01, 4.0]]. This final dataset therefore consists of 600 "strongly sentiment-related" words. Note that these samples were only used as part of the training set and not in the test set since these do not reflect our goal of detecting sentiment in a complete sentence.

Therefore, a combined dataset consisting of 1164 manually annotated user utterances from human-machine dialogues along with a dataset containing 600 sentiment words was used, for a total of 1764 samples.

\subsection{Testing Existing Systems}
\label{section3.2}

In order to detect the user’s sentiment precisely, this paper tested five different systems. These five systems are Naive Bayes classifier, Stanford CoreNLP, affective lexicon by Finn (AFINN), NLTK Vader and Random Forest classifier. \\
\indent \textbf{Random Forest classifier:-}
This system uses NLTK toolkit with random forest classifier was written in python script. This system is downloaded from Kaggle's competition. Originally, this system specialised in movie reviews sentiment analysis, but this paper uses it for testing Alana dataset. The random forest classifier is trained on 70\% of Alana data and test it on the rest 30\%. This system technique is creating a bag of words by utilising a tokenization process then creating the feature vectors in order to train the classifier using these vectors as input. \\
\indent \textbf{Affective lexicon by Finn (AFINN):-}
This system was written by using JavaScript (NodeJS). In general, this system is a wordlist-based system. In other words, This system contains a list of words. Each of these words has a value between -5 (negative) and 5 (positive). In order to calculate the sentiment value of the sentence, AFINN sums up all sentence’s words values and divide this result by a number of all words in this sentence. \\
\indent \textbf{NLTK Vader (Valence Aware Dictionary and Sentiment Reasoner):-}
This system is lexicon and rule-based system. The best domain for this system is the social media domain \citet{gilbert2014vader}. In general, this system gives the utterance a sentiment value between -1 as very negative and 1 as very positive \citet{lin2018sentiment}. \\
\indent \textbf{Stanford CoreNLP:-} Is coded by Java programming language, trained on movie reviews using Recursive Neural Network RNN. Stanford classifies the sentiment by understanding the meaning and the order of the words in a given sentence and handle the negation problem \citet{socher2013recursive}.\\
\indent \textbf{Naive Bayes:-} A sentiment classifier based on the Naive Bayes algorithm. It utilised the Bayes rule to calculate $P(label|features)$ in terms of $P(label)$ and $P(features|label)$.\\
\indent The experiments present that the random forest has the best accuracy among the other systems which this paper tested. This table shows the results for (-- , -, 0, +, ++) annotation categories.\\

\begin{table}[!ht]
\label{my-label1}
\begin{tabular}{|c|c|c|}
\hline
\textbf{System name} & \textbf{Acc} & \textbf{Alana Dataset use}      \\ \hline
Random Forest & { \textbf{ 69.1\%}}      & Training on 70\%     \\ 
                     &                   & Testing on 30\%      \\ 
                     &                   & with 25 trees        \\ \hline
AFINN                & 55.4\%            & Testing on all       \\ \hline
NLTK Vader           & 55.1\%            & Testing on all       \\ \hline
Stanford CoreNLP    & 53.6\%            & Training on 70\%     \\ 
                     &                   &  Testing on 30\%     \\ \hline
Naive Bayes          & 49.8\%            & Testing on all       \\ \hline
\end{tabular}
\caption{Systems Tested and Accuracy}
\end{table}

\indent \textbf{Experiment on random forest model:-}
Once the random forest has been chosen as best system, many improvements have been tried including: tuning the random forest hyper-parameters like increasing the number of the trees from 25 to 2000;
training on twitter dataset and test on Alana dataset; and combining Vader dataset (list of sentiment words) with Alana dataset. Table 2 shows the results.

\begin{table}[!ht]
\centering

\label{my-label2}
\begin{tabular}{|c|c|}
\hline
\textbf{Task}                   & \textbf{Accuracy} \\ \hline
Trained on 2881 twits,          & 49.5\%            \\ 
Tested on all Alana dataset     &                   \\ \hline
Increasing the number of the trees     & 70.2\%            \\ 
from 25 to 2000                 &                   \\ \hline
Combining Vader dataset         &                   \\ 
(600 sentiments words)          &\textbf{ 71.1\%}            \\ 
with Alana dataset, 25 trees    &                   \\ \hline
\end{tabular}
\caption{Random Forest Model Experiments}
\end{table}

\section{Alana Implementation}
Sentiment analysis is complex problem to solve, yet it can have a great impact on communication flow. Sentiment can be context-dependant, such as ~sarcasm~, some utterances can have strong vocal cues showing irony or sarcasm and people may use ~''workarounds''~ to express their feelings in indirect ways. Feelings are also very complex by themselves, and no human is capable to perfectly categorise them, let alone a machine.

\subsection{Alana system description}
\subsubsection{General introduction}
The Alana system is a chat-bot ~created/conceived~ by Heriot-Watt University's ~Natural Language Processing group~ (Edinburgh, Scotland). It participated in Amazon's Alexa 2017 Challenge and reached the finals, out of 15 competitors. 
The goal of this competition is to present the most capable open-domain dialogue system. Several iterative improvements have been thought of in order to enhance the performance of the system.
This paper describes an attempt at improving its sentiment analysis capabilities. In the following sections, a short technical introduction of the system followed by the implementation of our work inside the Alana system.

\subsubsection{Technical introduction}
Alana is a multi-agent system composed of multiple concurrently running "bots".  These "bots" are independent agents, built using various technologies, some of which use machine learning techniques while other use more scripted methods.
Most "bots" are specialised in a specific field such as news retrieval, giving facts or telling jokes.

\begin{figure}[h]
\centering
    \includegraphics[width=0.48\textwidth]{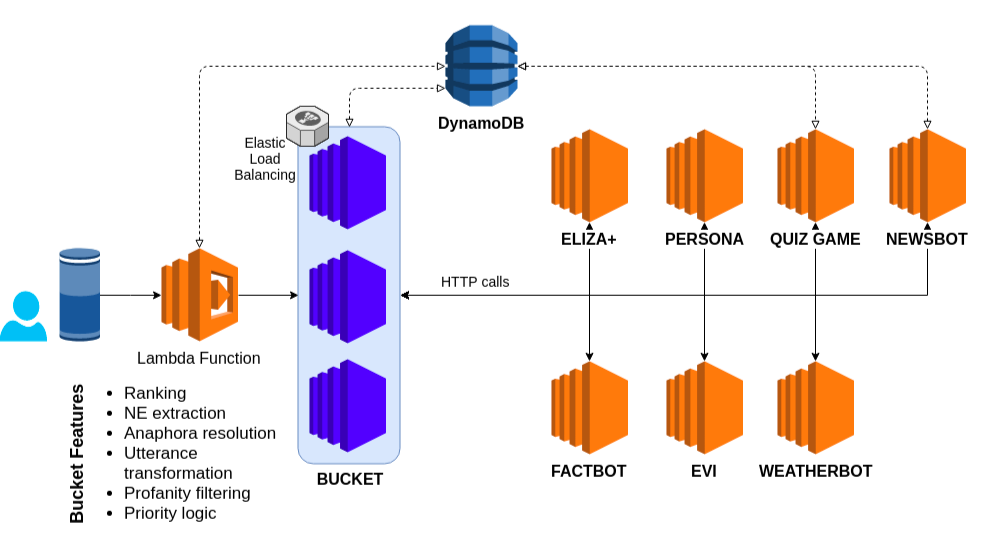}
        \caption[Alana System Overview]{Showing how Alana System is built, visualise what are the Bucket feature and have an overview of all the Chat-Bots}
    \label{fig:Alana System Overview}
\end{figure}

During a conversation, bots are queried after every user utterance and the final output is determined by a separate component called "Bucket".
This component is responsible for the general system orchestration, such as pre-processing input data, collecting bots output as well as producing the final system output.

\subsection{Implementation in Alana system}
\subsubsection{Attempt of implementation}
As a multi-agent system, Alana has a "bot" which was -partly- in charge of the conversation around sentiments. It was discovered later that this bot, called "Persona", wasn't using any proper sentiment analyser, but instead relies on some flexible AIML utterances that were already hard coded to create the impression of sentiment detection.
So, the first thought was to work around this bot, firstly by implementing into it into the Machine Learning Model and maybe by also improving how it talks through some added AIML.
It was discovered that a one to one implementation is not necessarily the best option, and it was preferable to implement this Sentiment Analyser on different level and in a flexible and responsive way. 

\subsubsection{Current implementation}
After some further research and some failed tests, a bucket implementation was settled on.
The idea was to add a general "prefixed" sentimentally polarized sentence before the vanilla system sentence if needed.

\begin{figure}[h]
\centering
    \includegraphics[width=0.48\textwidth]{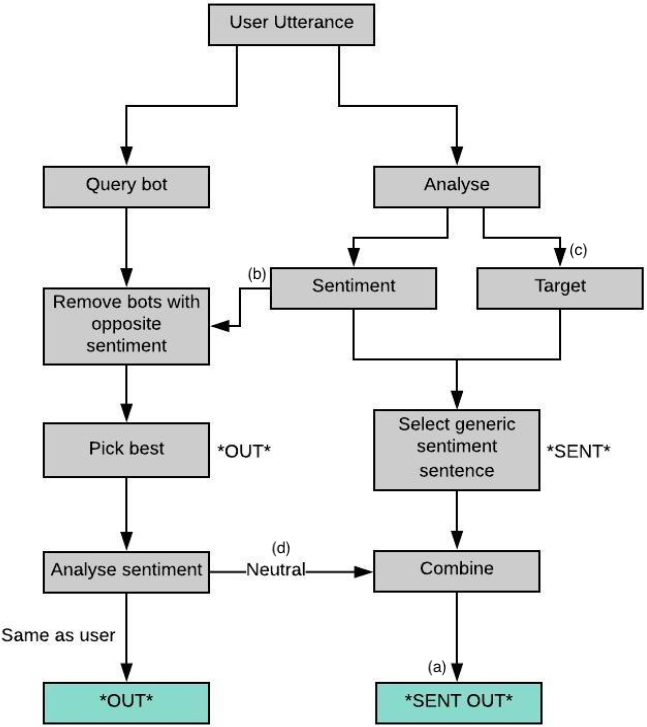}
        \caption[Alana Implementation Scheme]{(a) Pre-fix the system output with a generic sentiment filled sentence. (b) Prevent bots with opposite sentiment from being selected. (c) Detection of user sentiment’s target (Alana vs other). (d) If output is already sentiment filled, do not add pre-fix.}
    \label{fig:Alana Implementation}
\end{figure}

The general Alana system gets the user utterance as an input. This user utterance is also used as an input for the system being created, and its sentiment polarity is analyzed using the previously described machine learning algorithm.
This creates a sentiment as an output, which is compared to the sentiment that the vanilla Alana system may detect.
The Machine Learning Model is run on each bot.
\\
\indent If our output polarity is opposed to one of the bots polarity, we prevent this bot from being triggered by kicking it; in order to avoid having as a final output a sentence including two opposite sentiment such as "That's sad, I am so glad for you, [...]".
\\
\indent If the output polarity is the same as the vanilla system output polarity, the system prevents sending an output, and keeps only the vanilla system one; in order to avoid having as a final output a sentence including two times such as "I am so happy, I am so glad for you, [...]".
\\
\indent Finally, because the Machine Learning Model was not powerful enough because more annotated data would lead to better results, some negation or specific cases were not properly detected. A simple solution of switching polarity to the opposite sentiment if negation is in the sentence was added to the Bucket python file.
Finally a word like dead/death/die were flagged as negative by the system, so while asking for information on "Death Note" or "Die Antword" there was an output with a negative prefixed sentiment part before the introduction of the output sentiment. It was chosen for now to not run the model on the weather bot, the news bot, and either the wiki bot until a better solution to this problem could be found.

\subsubsection{Alternative implementations}
Detection of a strong feeling, either positive or negative is likely to drive a human-human discussion away from the topic. This means that the result of the sentiment analyser module should be able to influence the output of every bot, or give precedence to a dedicated bot capable to deal with this scenario. The first case is practically complex given the Alana system; every bot is considered and has black boxes in the context of our project, modifying each of them is deemed difficult. The second way would be a suitable solution but is likely to reduce the coherence the whole system has; i.e. transitioning from a phase ''handing sentiment'' to ''talking about any other topic''. Moreover, considering the time limitations of this project, the idea of building an additional bot could not be considered.

\subsubsection{Improvements to the current implementation}
Basically the first idea would be to find a better Machine Learning Model or add some better rules but a lot of systems were tested before choosing and building the current one.
\\
\indent Obviously, the most important thing to improve, would be to annotate a considerable amount of data, or improve the annotation scheme and method. The whole model is based on the data, and the better the data is, the better the system will act.
\\
\indent It was also thought to improve our model by bringing it under a different angle, with a Sentiment Analyser that can understand proper sentiment and not only their polarity. But then annotate of the data would need to accordingly change.The annotation scheme could for example be modified to : "pleased, disgusted, amused, intrigued" or something similar. 
\\
\indent Finally, it was also found that another system could be worked on to detect sarcasm. It was discovered that on some blogging platforms such as Reddit, people use "/s" or "/S" to flag sarcasm in their sentence. Indeed, while sarcasm is something that can be heard or seen but not read, using sentences from those platforms may help to work on a Sarcasm Detector. Nevertheless while machine learning can be done on some subreddit sentences, the issue at hand is that those are not part of dialogue utterances and may not be useful for the Sentiment Analyser. It then would need differentiation of this tool from the currently developed system with first analysis of sarcasm being done before analysing its sentiments. It is believed that a Sarcasm Detector could be something really useful for AI in general.

\section{Evaluation Method}

Regarding the user evaluation of the system, two different chat-bots were created and uploaded on Telegram: Susan, which was the current version of the Alana chat-bot and Rob, which was Susan equipped with the sentiment prediction model that was developed for this project. The users were instructed to chat with both bots and, afterwards, complete a survey form using their unique chat identification number. The users were asked to provide feedback on whether they felt each chat-bot was able to understand their feeling. In addition, users were also asked to provide a rating, on a scale of 0 to 5, on their general experience regarding the conversation with each bot.

\section{Results}

Throughout an evaluation period of $22$ days, $26$ people conversed with Susan and Rob. Their feedback is as follows:

\begin{itemize}
\item $23.1$\% of users feel that Susan was able to grasp their feeling, whereas
\item $61.5$\% of users feel that Rob was able to grasp their feeling.
\end{itemize}

This proves that there exists a significant improvement in the ability of the system to recognise sentiment. However, the exact percentage of improvement cannot be derived from these numbers. 

To answer this question, the user rating states that the average user score for Susan was $2.34$ out of $5$, whereas for Rob it was $2.84$. Such difference might seem unimportant, but a more careful glance shows a significant average of $21.37$\% increase to the user’s engagement with Rob, compared to Susan. 

\begin{figure}[h]
\centering
    \includegraphics[width=0.48\textwidth]{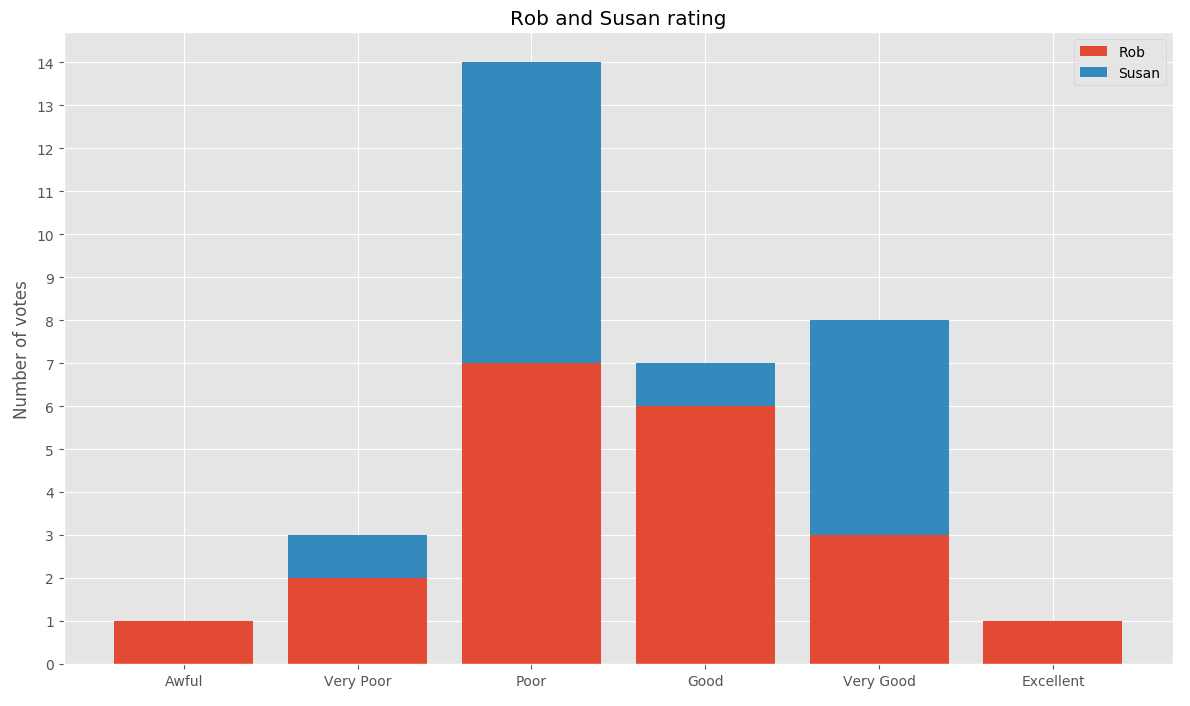}
    \label{fig:Susan Vs Rob}
\end{figure}

Despite the small size of the group that conversed with the chat-bots and completed the survey, the existence of an improvement in detecting sentiment is clear. What is not clearly shown up to this point is the exact percentage of this improvement. A larger user group would have smoothened the distribution, providing numbers with higher confidence. 

Another factor that might have influenced the evaluation process is the following: In order for a user to realise whether the system was able to grasp their feeling, an appropriate response has to be provided by the system just after the feeling's expression, as verification. In Rob's case, the system's response to the user was very generic, a decision made previously to simplify the implementation process and make the sentiment predictor independent from the rest of the system. As a result, the quality of the responses produced by the system was not optimal and, therefore, there have might have been cases where the system was able to detect a sentiment, but the system's response was inappropriate to the sentiment detected. Therefore, the user received the information that the system was unable to understand their emotional state and provided a low rating. Should the goal here be to increased user engagement and not to improve sentiment detection, the argument to be made is that, using the same sentiment prediction model and higher quality system responses would have yielded even better results and simulated a more engaging conversational environment.

\section{Conclusion and Future Work}

To allow users to interact with a conversation agent socially or more natural, the agent must be able to express its feeling and thoughts back to the user \citep{breazeal1999build}. However, as conversational agent such as Alana has no facial feature, the only output that Alana provides is a sound (in Alexa system) or a text message (in the current implementation). Thus, the only mean available for Alana to express its understanding of user utterance by a response that includes its thoughts and feelings toward user utterance. From the experiment in \autoref{section3.2}, the \textbf{Random Forest Model} yields the highest accuracy (69.1\%) among other models tested. So, to understand user feelings or emotions toward the topic of discussion, it was decided to analyse user's utterance using a \textbf{Random Forest Model} (25 trees)  trained on the interaction between user and Alana combined with the dataset from NLTK Vader. 

After the implementation of the \textbf{Random Forest Model} into Alana system and renaming it to \textbf{'Rob'} for testing purposes, an evaluation is run to compare Alana (a Alexa prize system) with Rob (our implementation) by letting users try to chat with 'Rob' and 'Susan' (renamed original Alana) without telling them which is the modified system. From the result of the evaluation, it is noticed that users report that 'Rob' tends to understand their feelings more than 'Susan' and they gave a better overall rating for 'Rob'. It can can concluded that to improve user experience when interacting with a conversational agent, the agent must be able to create a belief that it can understand user's thoughts and feelings by using sentiment analysis to determine user's thoughts and feelings.

Future research will be focused on the annotation quantity and quality. From the result in \autoref{section3.2}, it is believed that if there is more annotated data with greater consistency a better sentiment classifier could be trained. Also, investigation on a sentiment analysis on user's voice will be done. Instead of doing a sentiment analysis on user's utterance in a form of text data, it is felt that the emotions and feelings are incorporated in user's voice. In human utterance generation, each part in a vocal box (larynx) will contract and expand according to an individuals current emotion which results in a angry, sad, happy, or a neutral tone \citep{ying2018characteristics}. 
A sentiment classifier can be created to detect emotions or feelings from a user's voice to compare with the result of the current implementation.

\section*{Acknowledgements}

We would like to offer our sincerest gratitudes to Professor Verena Rieser of Heriot-Watt University for supervising the project and Xinnuo Xu of Heriot-Watt University for being an invaluable support as a research assistant that had previously worked on the Alana System. We would also like to thank our colleagues and peers at Heriot-Watt who participated in the evaluation stage of the project.

\bibliography{acl2018}

\begin{thebibliography}{12}
\expandafter\ifx\csname natexlab\endcsname\relax\def\natexlab#1{#1}\fi

\bibitem[{Appel et~al.(2016)Appel, Chiclana, Carter, and Fujita}]{APPEL2016110}
Orestes Appel, Francisco Chiclana, Jenny Carter, and Hamido Fujita. 2016.
\newblock \href {https://doi.org/https://doi.org/10.1016/j.knosys.2016.05.040}
  {A hybrid approach to the sentiment analysis problem at the sentence level}.
\newblock \emph{Knowledge-Based Systems}, 108:110 -- 124.
\newblock New Avenues in Knowledge Bases for Natural Language Processing.

\bibitem[{Breazeal and Scassellati(1999)}]{breazeal1999build}
Cynthia Breazeal and Brian Scassellati. 1999.
\newblock How to build robots that make friends and influence people.
\newblock In \emph{Intelligent Robots and Systems, 1999. IROS'99. Proceedings.
  1999 IEEE/RSJ International Conference on}, volume~2, pages 858--863. IEEE.

\bibitem[{Gilbert(2014)}]{gilbert2014vader}
CJ~Hutto~Eric Gilbert. 2014.
\newblock Vader: A parsimonious rule-based model for sentiment analysis of
  social media text.
\newblock In \emph{Eighth International Conference on Weblogs and Social Media
  (ICWSM-14). Available at (20/04/16) http://comp. social. gatech.
  edu/papers/icwsm14. vader. hutto. pdf}.

\bibitem[{Katz et~al.(2015)Katz, Ofek, and Shapira}]{KATZ2015162}
Gilad Katz, Nir Ofek, and Bracha Shapira. 2015.
\newblock \href {https://doi.org/https://doi.org/10.1016/j.knosys.2015.04.009}
  {Consent: Context-based sentiment analysis}.
\newblock \emph{Knowledge-Based Systems}, 84:162 -- 178.

\bibitem[{Lin et~al.(2018)Lin, Zampetti, Bavota, Di~Penta, Lanza, and
  Oliveto}]{lin2018sentiment}
Bin Lin, Fiorella Zampetti, Gabriele Bavota, Massimiliano Di~Penta, Michele
  Lanza, and Rocco Oliveto. 2018.
\newblock Sentiment analysis for so ware engineering: How far can we go?

\bibitem[{Papaioannou et~al.(2017)Papaioannou, Curry, Part, Shalyminov, Xu, Yu,
  Dušek, Rieser, and Lemon}]{PapaioannouIoannis2017AEMw}
Ioannis Papaioannou, Amanda~Cercas Curry, Jose~L. Part, Igor Shalyminov, Xinnuo
  Xu, Yanchao Yu, Ondřej Dušek, Verena Rieser, and Oliver Lemon. 2017.
\newblock \href
  {https://s3.amazonaws.com/alexaprize/2017/technical-article/alana.pdf}
  {Alana: Social dialogue using an ensemble model and a ranker trained on user
  feedback}.
\newblock In \emph{1st Proceedings of Alexa Prize (Alexa Prize 2017)}, page~10.

\bibitem[{Pozzi et~al.(2017{\natexlab{a}})Pozzi, Fersini, Messina, and
  Liu}]{Pozzi20171}
Federico~Alberto Pozzi, Elisabetta Fersini, Enza Messina, and Bing Liu.
  2017{\natexlab{a}}.
\newblock \href
  {https://doi.org/https://doi.org/10.1016/B978-0-12-804412-4.00001-2} {Chapter
  1 - challenges of sentiment analysis in social networks: An overview}.
\newblock In Federico~Alberto Pozzi, Elisabetta Fersini, Enza Messina, and Bing
  Liu, editors, \emph{Sentiment Analysis in Social Networks}, pages 1 -- 11.
  Morgan Kaufmann, Boston.

\bibitem[{Pozzi et~al.(2017{\natexlab{b}})Pozzi, Fersini, Messina, and
  Liu}]{Pozzi2017239}
Federico~Alberto Pozzi, Elisabetta Fersini, Enza Messina, and Bing Liu.
  2017{\natexlab{b}}.
\newblock \href
  {https://doi.org/https://doi.org/10.1016/B978-0-12-804412-4.00016-4} {Chapter
  16 - conclusion and future directions}.
\newblock In Federico~Alberto Pozzi, Elisabetta Fersini, Enza Messina, and Bing
  Liu, editors, \emph{Sentiment Analysis in Social Networks}, pages 239 -- 241.
  Morgan Kaufmann, Boston.

\bibitem[{Ram et~al.(2017)Ram, Prasad, Khatri, Venkatesh, Gabriel, Liu, Nunn,
  Hedayatnia, Cheng, Nagar, King, Bland, Wartick, Pan, Song, Jayadevan, Hwang,
  and Pettigrue}]{RamAshwin2018CATS}
Ashwin Ram, Rohit Prasad, Chandra Khatri, Anu Venkatesh, Raefer Gabriel, Qing
  Liu, Jeff Nunn, Behnam Hedayatnia, Ming Cheng, Ashish Nagar, Eric King, Kate
  Bland, Amanda Wartick, Yi~Pan, Han Song, Sk~Jayadevan, Gene Hwang, and Art
  Pettigrue. 2017.
\newblock \href
  {https://s3.amazonaws.com/alexaprize/2017/technical-article/alexaprize.pdf}
  {Conversational ai: The science behind the alexa prize}.
\newblock In \emph{1st Proceedings of Alexa Prize (Alexa Prize 2017)}, page~18.

\bibitem[{Singh et~al.(2016)Singh, Singh, and Singh}]{Singh2016}
Jaspreet Singh, Gurvinder Singh, and Rajinder Singh. 2016.
\newblock \href {https://doi.org/10.1007/s40012-016-0107-y} {A review of
  sentiment analysis techniques for opinionated web text}.
\newblock \emph{CSI Transactions on ICT}, 4(2):241--247.

\bibitem[{Socher et~al.(2013)Socher, Perelygin, Wu, Chuang, Manning, Ng, and
  Potts}]{socher2013recursive}
Richard Socher, Alex Perelygin, Jean Wu, Jason Chuang, Christopher~D Manning,
  Andrew Ng, and Christopher Potts. 2013.
\newblock Recursive deep models for semantic compositionality over a sentiment
  treebank.
\newblock In \emph{Proceedings of the 2013 conference on empirical methods in
  natural language processing}, pages 1631--1642.

\bibitem[{Ying and Xue-Ying(2018)}]{ying2018characteristics}
Sun Ying and Zhang Xue-Ying. 2018.
\newblock Characteristics of human auditory model based on compensation of
  glottal features in speech emotion recognition.
\newblock \emph{Future Generation Computer Systems}, 81:291--296.

\end{thebibliography}
\bibliographystyle{acl_natbib}

\appendix


\section{Supplemental Material}

\begin{table*}[]
\centering
\label{my-label3}
\begin{tabular}{|l|l|l|l|}
\hline
\textbf{}            & \textbf{\# of categories} & \textbf{Accuracy} & \textbf{Notes}              \\ \hline
\textbf{Naïve Bayes} & 5                         & 49.8\%            & 70\% training, 30\% testing \\ \hline
\textbf{Stanford CoreNLP}    & 5                         & 53.6\%            & Tested on all Alana dataset \\ \hline
\end{tabular}
\caption{Other System Results}
\end{table*}

\begin{table*}[]
\centering
\label{my-label4}
\begin{tabular}{|c|c|c|c|c|c|c|}
\hline
\textbf{\#} & \textbf{\begin{tabular}[c]{@{}c@{}}\# of \\ categories\end{tabular}} & \textbf{\begin{tabular}[c]{@{}c@{}}\# of \\ trees\end{tabular}} & \textbf{Precision} & \textbf{Recall} & \textbf{F-score} & \textbf{Alana Dataset use}                                                                                   \\ \hline
1           & 3                                                                    & 25                                                              & 76.4\%             & 76.2\%          & 75.8\%           & \begin{tabular}[c]{@{}c@{}}70\% training, \\ 30\% testing\end{tabular}                           \\ \hline
2           & 3                                                                    & 50                                                              & 76.9\%             & 76.5\%          & 76\%             & \begin{tabular}[c]{@{}c@{}}70\% training, \\ 30\% testing\end{tabular}                           \\ \hline
3           & 3                                                                    & 100                                                             & 77.2\%             & 76.8\%          & 76.3\%           & \begin{tabular}[c]{@{}c@{}}70\% training, \\ 30\% testing\end{tabular}                           \\ \hline
4           & 3                                                                    & 1000                                                            & 77.2\%             & 76.8\%          & 76.3\%           & \begin{tabular}[c]{@{}c@{}}70\% training, \\ 30\% testing\end{tabular}                           \\ \hline
5           & 3                                                                    & 2000                                                            & 77.5\%             & 77.1\%          & 76.7\%           & \begin{tabular}[c]{@{}c@{}}70\% training, \\ 30\% testing\end{tabular}                           \\ \hline
6           & 5                                                                    & 25                                                              & 68.9\%             & 69.1\%          & 68\%             & \begin{tabular}[c]{@{}c@{}}70\% training, \\ 30\% testing\end{tabular}                           \\ \hline
7           & 5                                                                    & 50                                                              & 67.4\%             & 68.5\%          & 67.8\%           & \begin{tabular}[c]{@{}c@{}}70\% training, \\ 30\% testing\end{tabular}                           \\ \hline
8           & 5                                                                    & 100                                                             & 68.5\%             & 68.2\%          & 67.7\%           & \begin{tabular}[c]{@{}c@{}}70\% training, \\ 30\% testing\end{tabular}                           \\ \hline
9           & 5                                                                    & 1000                                                            & 68.8\%             & 69.4\%          & 68.6\%           & \begin{tabular}[c]{@{}c@{}}70\% training, \\ 30\% testing\end{tabular}                           \\ \hline
10          & 5                                                                    & 2000                                                            & 68.4\%             & 70.2\%          & 69.3\%           & \begin{tabular}[c]{@{}c@{}}70\% training, \\ 30\% testing\end{tabular}                           \\ \hline
11          & 5                                                                    & 25                                                              & 65.2\%             & 66.9\%          & 64.5\%           & \begin{tabular}[c]{@{}c@{}}70\% training, \\ 30\% testing \\ without ambiguous flag\end{tabular} \\ \hline
12          & 5                                                                    & 25                                                              & 70.4\%             & 71.1\%          & 70.5\%           & \begin{tabular}[c]{@{}c@{}}70\% training + \\ Vader dataset, \\ 30\% testing\end{tabular}        \\ \hline
13          & 3                                                                    & 100                                                             & 68.7\%             & 49.5\%          & 37.8\%           & \begin{tabular}[c]{@{}c@{}}Trained on 2881 twits,\\ Tested on all Alana dataset\end{tabular}       \\ \hline
\end{tabular}
\caption{Random Forest Results}
\end{table*}

\begin{table*}[]
\centering
\label{my-label5}
\begin{tabular}{|l|l|l|l|l|}
\hline
\textbf{Annotation}    & \textbf{Precision} & \textbf{Recall} & \textbf{F-score} & \textbf{Support} \\ \hline
Very negative & 40\%      & 40\%   & 40\%    & 10      \\ \hline
Negative      & 50\%      & 23\%   & 31\%    & 22      \\ \hline
Neutral       & 81\%      & 82\%   & 82\%    & 169     \\ \hline
Positive      & 69\%      & 74\%   & 71\%    & 122     \\ \hline
Very positive & 39\%      & 41\%   & 40\%    & 27      \\ \hline
Total         & 71\%      & 71\%   & 70\%    & 350     \\ \hline
\end{tabular}
\caption{Classification Report for 5 Categories and 25 Trees (Alana and Vader Dataset}
\end{table*}

\begin{table*}[]
\centering
\label{my-label6}
\begin{tabular}{|l|l|l|l|l|l|}
\hline
\textbf{\#} & \textbf{\# of categories} & \textbf{Precision} & \textbf{Recall} & \textbf{F-score} & \textbf{Notes}              \\ \hline
1           & 3                         & 71.9               & 68.9\%          & 68.8\%           & Tested on all Alana dataset \\ \hline
2           & 5                         & 55.5\%             & 55.4\%          & 53.4\%           & Tested on all Alana dataset \\ \hline
\end{tabular}
\caption{AFINN Results}
\end{table*}

\begin{table*}[]
\centering
\label{my-label7}
\begin{tabular}{|l|l|l|l|l|l|}
\hline
\textbf{\#} & \textbf{\# of categories} & \textbf{Precision} & \textbf{Recall} & \textbf{F-score} & \textbf{Notes}              \\ \hline
1           & 3                         & 66.1\%             & 58.7\%          & 55.8\%           & Tested on all Alana dataset \\ \hline
2           & 5                         & 61.5\%             & 55.1\%          & 53\%             & Tested on all Alana dataset \\ \hline
\end{tabular}
\caption{NLTK Vader Results}
\end{table*}

\end{document}